\documentclass[letterpaper, 10 pt, conference]{ieeeconf}  
\usepackage{amsmath}
\usepackage{tensor}
\usepackage{bm}
\usepackage{amssymb}
\usepackage{graphicx}
\usepackage{caption}
\usepackage{color}
\usepackage{midfloat}
\usepackage{float}
\usepackage{rotating}
\usepackage{multirow}
\usepackage{hhline}
\usepackage{booktabs}
\usepackage{makecell}
\usepackage{longtable}
\usepackage{array}
\usepackage{cite}
\usepackage{subscript}
\usepackage{paralist}
\usepackage{multicol}
\usepackage{chngcntr}
\usepackage{arydshln,leftidx,mathtools}

\IEEEoverridecommandlockouts                              

\title{\LARGE \bf
        Adaptive and Multi-object Grasping via Deformable Origami Modules
}

 	\author{
        Peiyi Wang*, Paul A. M. Lefeuvre*, Shangwei Zou, Zhenwei Ni, Daniela Rus and Cecilia Laschi
		\thanks{This work funded by the National Research Foundation (NRF), Prime Minister’s Office, Singapore under its Campus for Research Excellence and Technological Enterprise (CREATE) programme. The Mens, Manus, and Machina (M3S) is an interdisciplinary research group (IRG) of the Singapore MIT Alliance for Research and Technology (SMART) centre. (\textit{Corresponding: Peiyi Wang}) (*: contribute equally to this work)}
        \thanks{Peiyi Wang, and Daniela Rus are with Singapore MIT Alliance for Research and Technology (SMART) centre, Singapore, Singapore}
        \thanks{Peiyi Wang, Paul A. M. Lefeuvre, Shangwei Zou, Zhenwei Ni and Cecilia Laschi are with Department of Mechanical Engineering, National University of Singapore, Singapore, Singapore}
        \thanks{Daniela Rus is with CSAIL, Massachusetts Institute of Technology, Cambridge, USA}
	}
	
\begin{document}
		
		\maketitle
		\thispagestyle{empty}
		\pagestyle{empty}

		\begin{abstract}
          Soft robotics gripper have shown great promise in handling fragile and geometrically complex objects. However, most existing solutions rely on bulky actuators, complex control strategies, or advanced tactile sensing to achieve stable and reliable grasping performance. In this work, we present a multi-finger hybrid gripper featuring passively deformable origami modules that generate constant force and torque output. Each finger composed of parallel origami modules is driven by a 1-DoF actuator mechanism, enabling passive shape adaptability and stable grasping force without active sensing or feedback control. More importantly, we demonstrate an interesting capability in simultaneous multi-object grasping, which allows stacked objects of varied shape and size to be picked, transported and placed independently at different states, significantly improving manipulation efficiency compared to single-object grasping. These results highlight the potential of origami-based compliant structures as scalable modules for adaptive, stable and efficient multi-object manipulation in domestic and industrial pick-and-place scenarios.

		\end{abstract}
		\section{Introduction}
                As robotics continues to expand beyond industrial automation into unstructured environments and daily tasks, there is a growing demand for efficient grippers that can handle objects with varying geometries and stiffness, and even multi-object grasping \cite{eom2024mogrip}.
                Anthropomorphic hands, with multi-finger geometry, represent the current pinnacle of dexterous manipulation \cite{li2024comprehensive, song2025overview, shadow-robot-2025}, as their multiple actuated fingers enable human-like coordinated control.
                However, this dexterity comes at the cost of substantial hardware and control complexity, requiring numerous actuators and precise coordination schemes that limit their scalability and practicality.
                In contrast, soft robotics has emerged as a promising alternative. By exploiting the inherent compliance of elastomeric materials, soft robotic systems \cite{laschi2016soft} embody mechanical intelligence, achieving adaptive interaction and safe contact without relying on complex perception or feedback control.
                Soft grippers are lightweight, inexpensive, and intrinsically safe, qualifying them for tasks such as marine sample collection \cite{sinatra2019ultragentle}, HASEL-based fruit grasping \cite{acome2018hydraulically}, and bio-inspired grasping in daily scenarios \cite{xie2020octopus}.
                
                

                Although existing soft grippers \cite{xie2020octopus, liang2025bio, glick2018soft} excel at conformal, form-fit grasps, they often lack the ability to regulate contact force, making them unsuitable for fragile or deformable objects.
                To address this issue, several sensorized soft grippers have been developed \cite{xie2023octopus, jin2020triboelectric, maruyama2013delicate, tawk2022force, mun2024multi}.
                For example, Maruyama et al. \cite{maruyama2013delicate} introduced a two-finger gripper with fluid-based fingertips that measure internal pressure for force estimation; however, its grasping capability is restricted to tip contact and cannot adapt to irregular shapes.
                Mun et al. \cite{mun2024multi} proposed a bellow-based design with optical force sensing, while Tawk et al. \cite{tawk2022force} integrated pneumatic sensors into rigid fingers.
                Despite these advancements, such systems remain sensor-dependent, bulky, and less compliant, limiting their generality and integration potential.
                
\begin{figure*}[!t]
            			\centering
            			\includegraphics[width=0.9\linewidth]{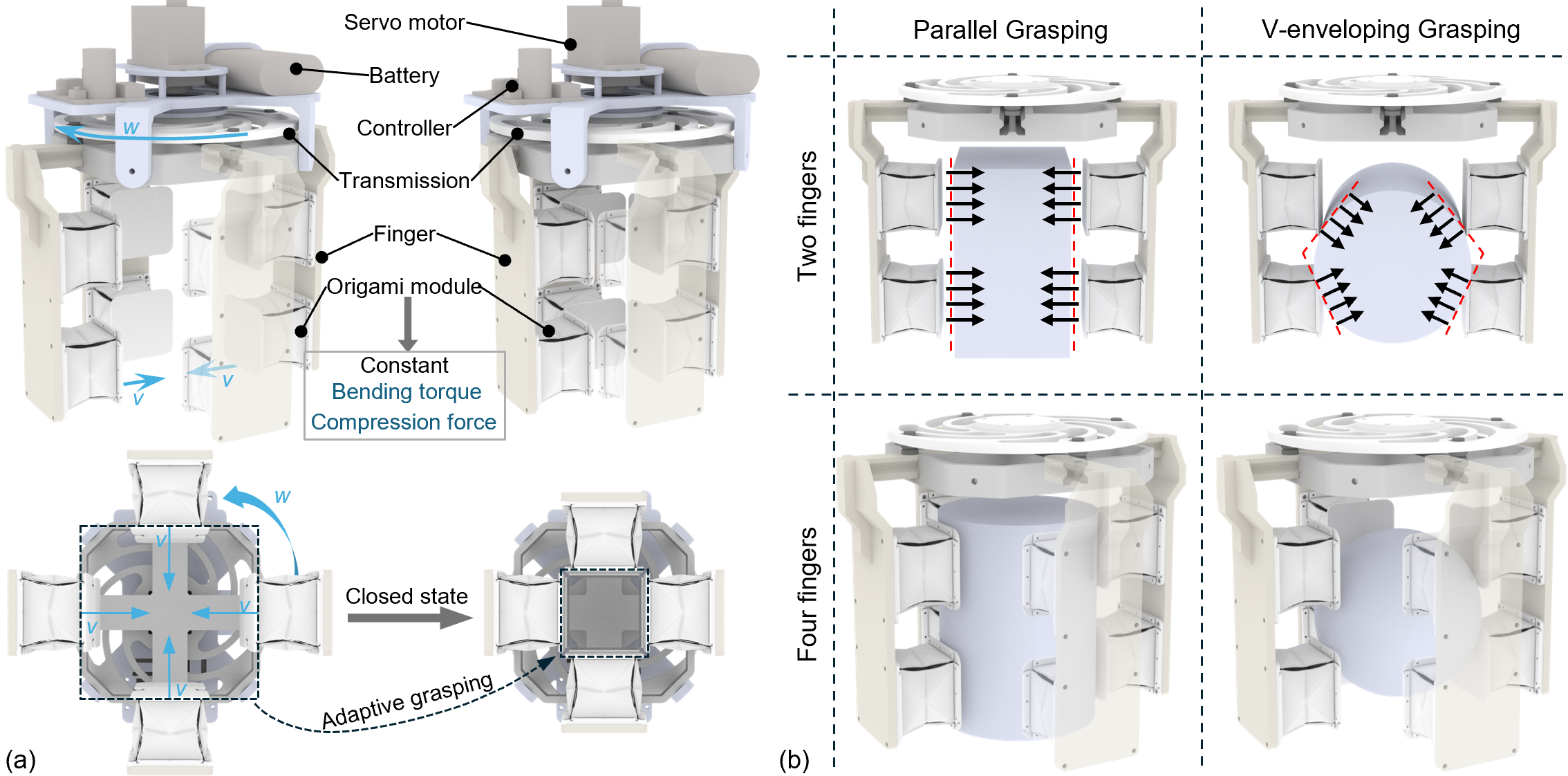}
            			\caption{System design and functionality of a multi-finger hybrid gripper with passively deformable origami modules. (a) The whole and bottom view of the design gripper in fully open and closed states. (b) Adaptive grasping such as parallel and v-enveloping grasping using two or four fingers.}
            			\label{Fig_ConceptDesign}
            		\end{figure*}

                In this work, we propose a multi-finger hybrid soft robotic gripper that employs passively deformable origami modules \cite{ni2025origami} as a means of mechanical intelligence for adaptive grasping.
                Each origami module, composed of compliant panels and hinge-like folds, generates a constant output force and torque within a defined deformation range, enabling passive force control \cite{maruyama2013delicate, wang2014constant} and shape adaptability \cite{lee2020shape}. Depending on the object geometry, the gripper can automatically transition between parallel and v-enveloping grasp configurations \cite{feix2015grasp}, achieving secure and stable contact without active sensing (Fig. \ref{Fig_ConceptDesign}).
                Moreover, by leveraging the constant-force property of the origami modules, the gripper can perform simultaneous multi-object grasping and sequential placement simply by adjusting grasp size, without additional control inputs.

                Origami-inspired mechanisms \cite{wheeler2016soft} offer a unique combination of precise motion control and inherent compliance, owing to their fold-based degree-of-freedom (DoF) and geometric constraints.
                Such designs have been widely explored for locomotion \cite{ze2022soft}, actuation \cite{jeong2018design}, and gripping \cite{chen2021soft, li2019vacuum, liu20213d}.
                However, the use of soft origami modules for both passive force regulation and multi-object manipulation remains largely unexplored.
                While Liu et al. \cite{liu2023hybrid} introduced a hybrid gripper with limited multi-object grasping capabilities, it lacks controllable or constant mechanics behavior, restricting its adaptability.

                This work advances the state of the art by demonstrating that deformable origami modules can serve as passive, tunable units for adaptive and multi-object grasping.
                Through systematic characterization and experiments, we validate that the proposed gripper maintains predictable constant mechanics response while stably manipulating single and multiple objects of diverse geometries and stiffness.
                Our design represents a sensor-free and mechanically intelligent approach toward simplifying the control of multi-finger robotic grippers for practical manipulation in complex environments.
                
        

        \section{System Design of Multi-finger Gripper}
                A parallel and hybrid soft gripper with multiple fingers, each with two origami-inspired modules, is designed for adaptive and multi-object grasping (shown in Fig. \ref{Fig_ConceptDesign}(a)). The 1-DoF actuation mechanism is driven by a servo motor through a radial expanding rail. The rotation of servo motor can be transmitted to linear motion of multiple fingers simultaneously. When the gripper attempts to grip an object, all fingers are simultaneously passively deformed and adapted to the object. As the deformable origami modules make contact, their origami panels deform with constant torque and force output due to its specific folding pattern \cite{ni2025origami}.

                Each origami module has four concentrically arranged origami panels. During compression, a consistent and stable force output is generated due to the structure mitigating inter-layer motion coupling and enabling independent layer deformation. Two modules are fixed vertically on each finger. The fingers provide structural support for each module to maintain a parallel grasp.
                
                The transmission disk drives finger motion with four spiral guides determined by $r(\theta)=54-\frac{25}{90}\theta,\{0^\circ\leq\theta\leq90^\circ\}$ (mm). At maximum grasp size, the servo is at 0°, which enables up to 78 mm opening between the origami modules. At minimum grasp size, the servo is at 90°. Both grasp sizes are shown in Fig. \ref{Fig_ConceptDesign}(a).

                The deformable quality of the origami modules allows them to passively adapt to grasping object geometry while nonetheless applying the same constant force. Regardless of the number of fingers, two principal grasp types can be achieved: parallel grasping with more cubic objects, and v-enveloping grasping for more spherical objects (shown in Fig. \ref{Fig_ConceptDesign}(b)). In the v-enveloping grasp, the passive compliance of the modules creates a circular wrapping around the target object which benefits from both force and form closure. Conversely, the parallel grasp only has force closure through the application of friction forces.
        
        \section{Origami Module} 
            \subsection{Material and Fabrication}
            The deformable origami module was fabricated in two different materials: TPU with a Shore hardness of 95A via 3D printing, and Smooth-Sil\texttrademark-950 Silicon via mold casting. These materials have the benefit of being inexpensive and easy to use. Each silicon panel was manufactured using injection molding. The soft panels and rigid PLA disks were assembled together, with each panel positioned at a radial distance of 15 mm. Rubber grip tape was applied to a rigid disk on one side to increase grasping adhesion.

            		\begin{figure}[!t]
            			\centering
            			\includegraphics[width=0.9\linewidth]{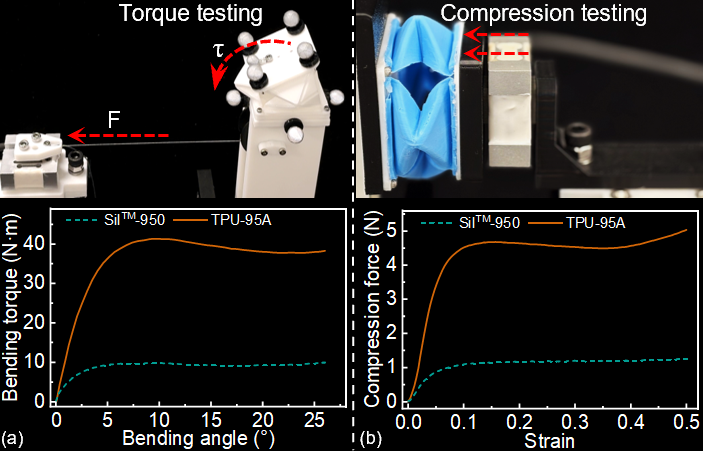}
            			\caption{ Experimental setup and measuring results for (a) bending and (b) compression force of the origami module. 
                        }
            			\label{Fig_CompressionTorque}
            		\end{figure}    
            
            \subsection{Force and Torque Tests}
            To evaluate each module's torque and compression characteristics, a 3-axis force gauge with accuracy 0.3 \% and range 0 - 20 N was affixed to a linear stage. As shown in Fig. \ref{Fig_CompressionTorque}(a), torque behavior was evaluated by pulling a thread attached to the origami module. The stepper motor applied torque by moving at a rate of 7.25 mm/s over 14.5 mm, then returned to its original position at the same speed. Several reflective markers were placed to accurately capture the bending angle of the module, after which the bending torque was calculated in data post-processing. After the initial loading phase, both materials maintained near-constant bending torques for the 5 - 25° range. The TPU-based module exerted 39 Nm of torque on average and stayed within a $\pm$5 \% range, higher than the silicon-based module (9.5 Nm average torque), but also with greater variance in values. This was likely due to differences in material properties. Comparatively, the silicone-based module had output a significantly more linear torque/angle relationship (within $\pm$3 \% deviation).

            Compression force was measured by pushing the force gauge directly into the module (Fig. \ref{Fig_CompressionTorque}(b)). The silicone-based module maintained a constant compression force output of 1 N, while the TPU-based module stayed within an acceptable range of 4.5 - 5 N over the 0.1 - 0.5 effective strain.
            
            
            
        \section{Experiments}

            We fabricate and assemble the gripper with two or four fingers according to different grasping tests and applications. To test gripper performance, we focused on evaluating its ability to handle tasks relevant to usage in a domestic setting. Three experiments were carried out, evaluating the gripper's performance for heavy objects, complex geometries, and multi-object grasping.

            		\begin{figure*}[!t]
            			\centering
            			\includegraphics[width=0.8\linewidth]{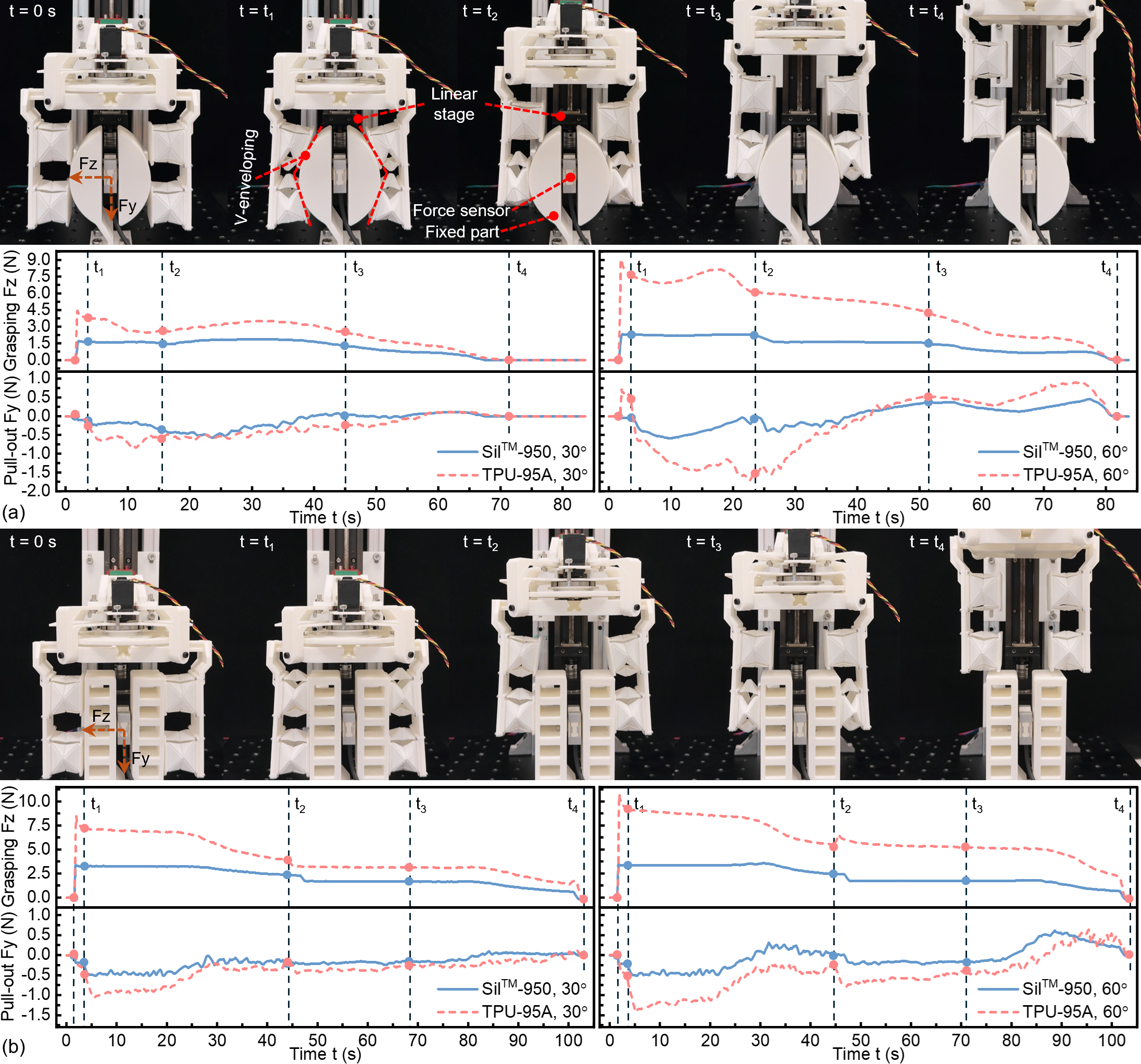}
            			\caption{Grasping and pull-out force tests under (a) v-enveloping grasp and (b) parallel grasp with experimental snapshots showing key grasping stages. The gripper fully grasps and the linear stage activates at t$_1$. At t$_2$, the top origami modules are fully pulled out. At t$_3$, the top modules have lost full contact and only the bottom modules are providing a grasping force. At t$_4$, the gripper has been fully pulled out. Constant force outputs occur in parallel grasping, for example, at time t$_3$.}
            			\label{Fig_GraspPullOutTest}
            		\end{figure*}
            
            \subsection{Grasping and Pull-out Force}
            To evaluate the grasping performance and functionality, both parallel and v-enveloping grasps were tested using a 3-axis force gauge in a linear stage platform. The sensor was placed within two symmetrical blocks, with one side fixed to the work surface to form a sensor probe. In each trial, after executing the grasp with the probe for approximately 2s, the stage lifted the gripper upward until full release. Each grasp configuration was tested at 30° and 60° servo angles using two materials (Sil\texttrademark-950 and TPU-95A). Testing results of grasping and pull-out force, as well as experimental snapshots of the key grasping stages, are shown in Fig. \ref{Fig_GraspPullOutTest}.

            Two curved blocks with radius 45.5 mm and width 29.4 mm were used for the v-enveloping grasping (Fig. \ref{Fig_GraspPullOutTest}(a)). Two cuboid blocks with dimensions 45.4$\times$100 mm were used for parallel grasping (Fig. \ref{Fig_GraspPullOutTest}(b)). Distinct grasping properties were observed between these two grasp modes. In the parallel grasp, where the origami modules undergo principally linear compression, both the TPU-based and silicone-based modules exhibit a nearly constant force output after approximately t\textsubscript{1} and t\textsubscript{2}, with a large range of constant output , confirming the constant-force characteristic of the origami structure. 

            In contrast, the v-enveloping grasp involves bending deformation of the origami modules, resulting in a nonlinear response rather than constant output. The silicone-based module maintaine moderate grasping forces (around 1.5 N) suitable for delicate objects, while the TPU-based modules provide higher initial forces (up to 7.5 N) but exhibited noticeable fluctuations, likely due to overfolding beyond the stable bending range or hysteresis of the TPU materials. The maximum pull-out force reaches 1.5 N for TPU-based modules at a 60° grasp, equivalent to a 3 N lifting capability in the two-finger configuration. These results highlight a trade-off between compliance and stability, where the parallel grasp mode offers predictable constant-force behavior, while the v-enveloping mode provides adaptive yet variable force for irregular shapes.

            \subsection{Multi-object Grasping and Efficiency Evaluation}
            The constant-output characteristics of the origami modules over a broad range of compression and bending enable the gripper to selectively grasp and release multiple objects simultaneously by simply adjusting the grasp size. This passive adaptability allows the gripper to maintain stable contact with one object while manipulating another, without any sensing or control feedback.
    
\begin{figure}[!t]
            			\centering
            			\includegraphics[width=0.9\linewidth]{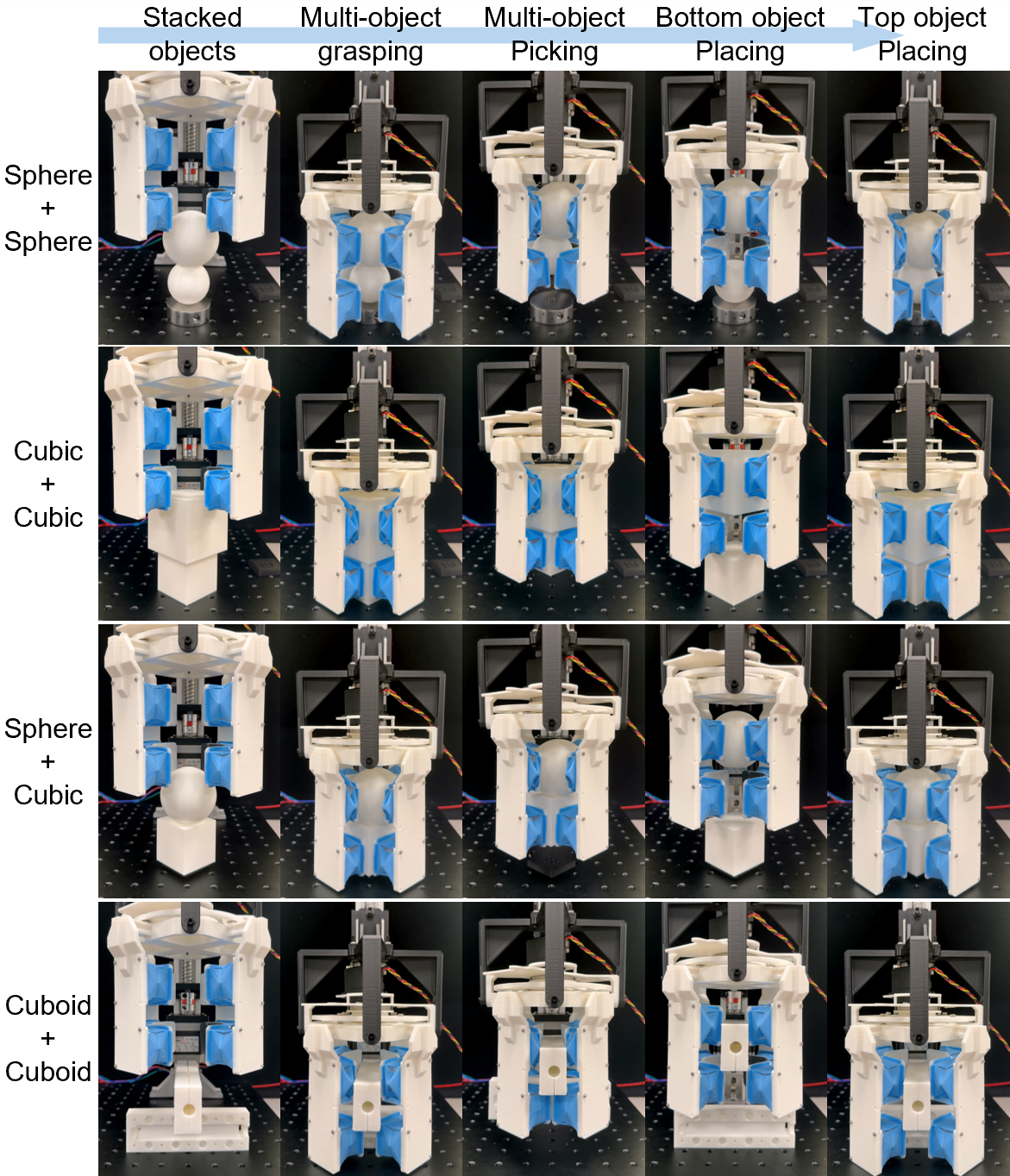}
            			\caption{
                        Multi-object grasping of stacked objects of various shapes and sizes, such as stacked spheres, cubes and cuboid blocks. The grasping process: pre-grasp position for stacked objects, multi-object grasping, multi-object picking, bottom object placing and top object placing.}
            			\label{Fig_MultiObjectGrasp}
            		\end{figure}

            
            To demonstrate this capability, pairs of simple geometries were vertically stacked and subjected to multi-object pick-and-place tests, as shown in Fig. \ref{Fig_MultiObjectGrasp}. The gripper began 60 mm above the objects and was subsequently lowered three times by a stepper motor moving at 10 mm/s: first, to pick up both objects with the smallest grasp size (90° servo angle), then to release the bottom object only with an intermediate grasp size (40°), finally to place the top object with the largest grasp size. Tests were conducted with four combinations: two spheres, two cubes, a sphere–cube pair, and two cuboids—covering both regular and long geometries. 
            As illustrated in Fig. \ref{Fig_MultiObjectGrasp}, the gripper successfully retained a stable grasp on the upper object while releasing the lower one, leveraging the deformable range of the origami modules to mechanically filter object size via 1-DoF mechanism. This demonstrates the feasibility of simultaneous multi-object manipulation purely through passive mechanical compliance.


\begin{figure}[!t]
            			\centering
            			\includegraphics[width=0.95\linewidth]{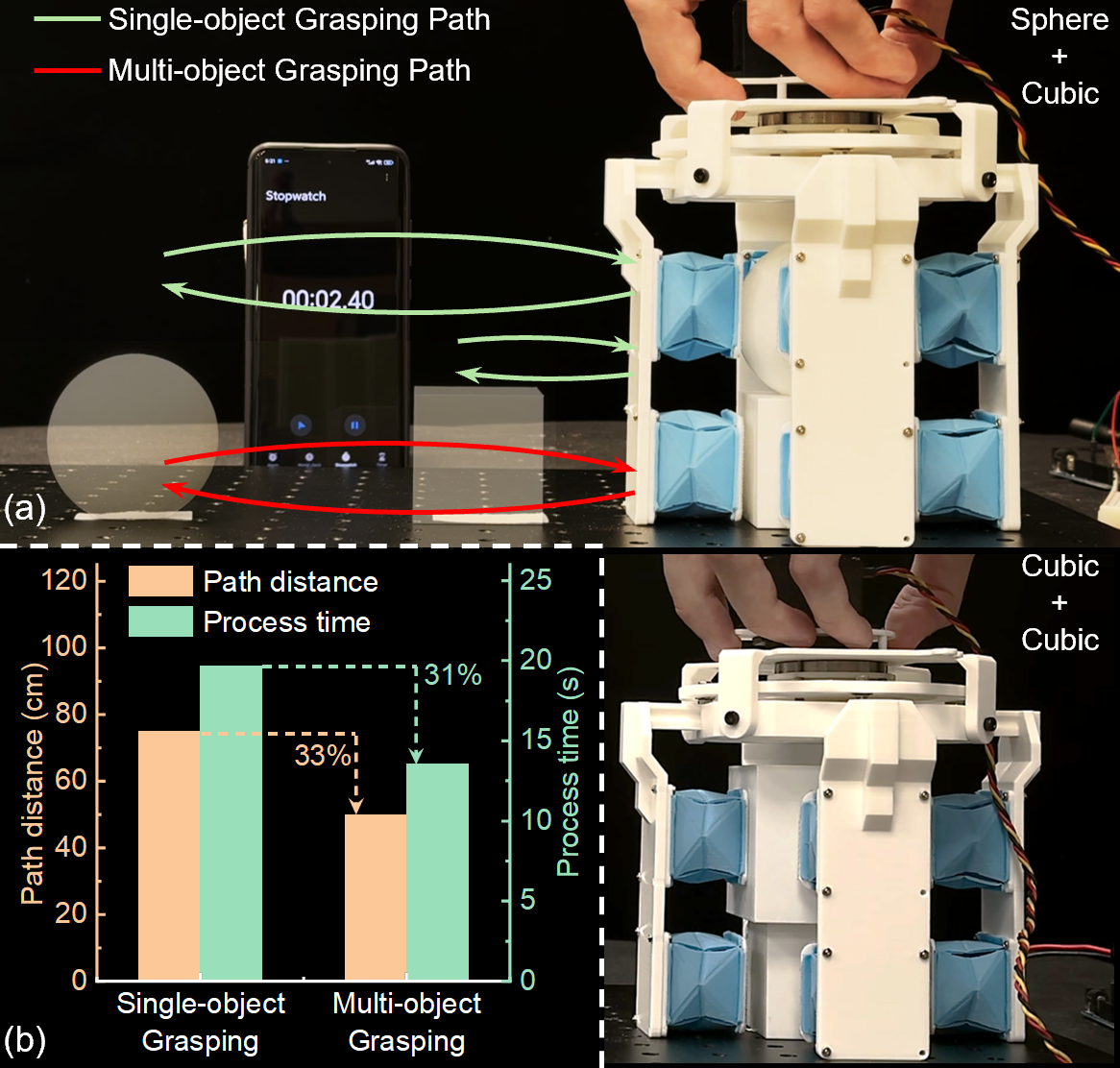}
            			\caption{Comparison of single-object and multi-object grasping. (a) Process and grasping path for picking and placing stacked objects. (b) Path distance and process time.}
            			\label{Fig_ComparisonTime}
            		\end{figure}

            To further evaluate its potential in practical manipulation tasks, a comparison was made between single-object sequential grasping and multi-object grasping, as shown in Fig. \ref{Fig_ComparisonTime}(a). In the single-object grasp, two individual pick-and-place cycles were required, while in the multi-object mode, both objects were handled within a single continuous motion. 
            Interestingly, the multi-object grasp eliminated one full back-and-forth trajectory, reducing the path distance by 33 \% and the overall process time by 31 \% (Fig. \ref{Fig_ComparisonTime}(b)). These results demonstrate that the proposed gripper not only enables simultaneous grasping but also improves manipulation efficiency without added control complexity, suggesting promising potential for industrial pick-and-place and logistics automation tasks that are repetitive or involve size-varied objects.
            
            \subsection{Grasping of Daily Objects with Complex Geometries}
\begin{figure}[!t]
                        \centering
                        \includegraphics[width=0.95\linewidth]{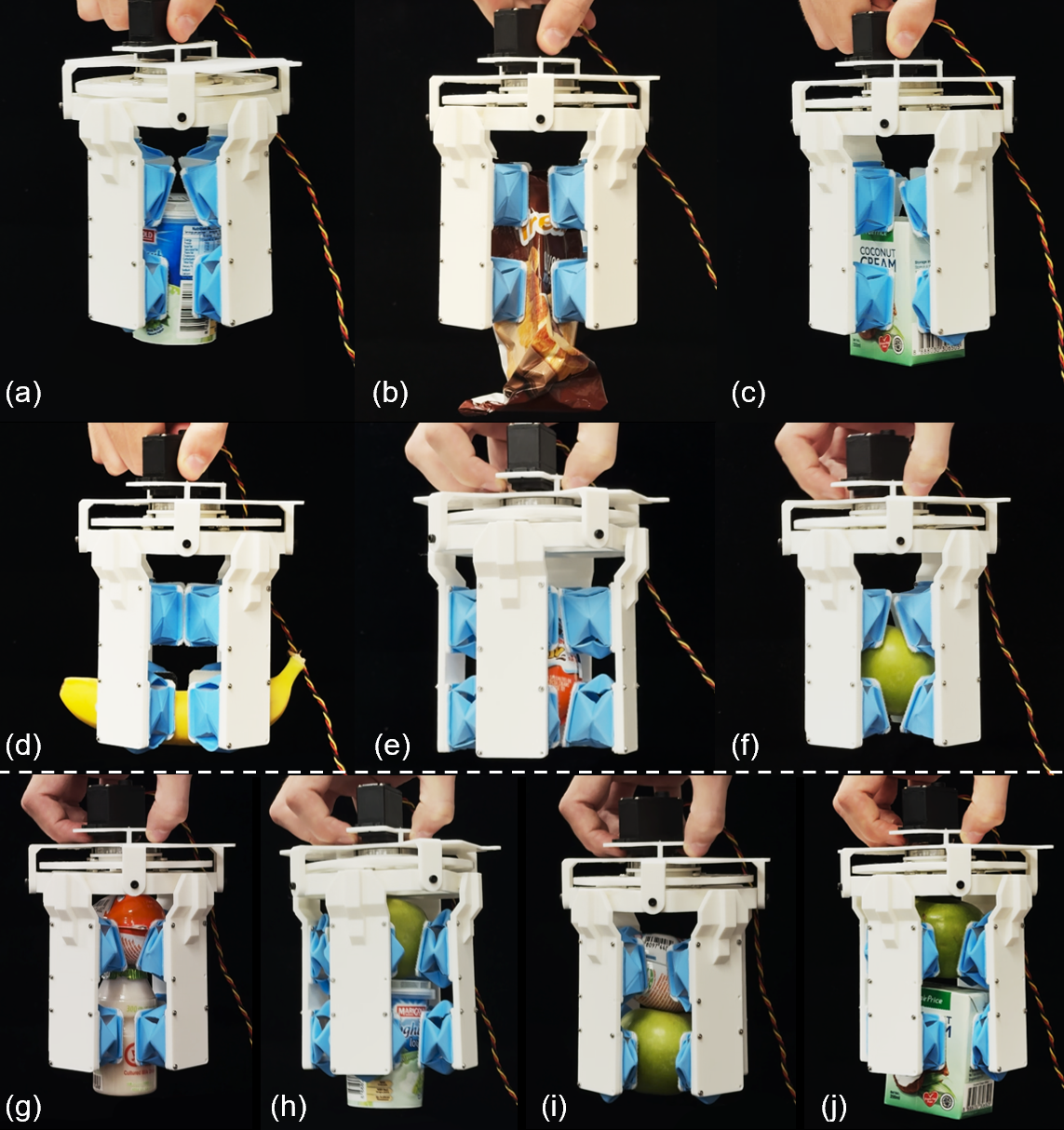}
                        \caption{Grasping experiments of daily uses. (a)-(f) Single-object grasping (a yogurt cup, a snack, a carton, a banana, an egg-shaped candy, an apple). (g)-(j) Multi-object grasping (egg with small drink, apple with yogurt cup, egg with apple, apple with carton).}
                        \label{Fig_ComplexGrasp}
                    \end{figure}

            To further verify the gripper's real-world applicability, a series of grasping tests were conducted on daily-use objects with diverse shapes, materials, and stiffness (shown in Fig. \ref{Fig_ComplexGrasp}). The origami modules passively adjust the grasp configuration to follow the grasped object geometry, applying the same grip force regardless of object size. As demonstrated in Fig. \ref{Fig_ComplexGrasp}(a–f), the gripper successfully handled objects including a yogurt cup, a snack bag, a carton, a banana, an egg-shaped candy, and an apple. For soft packaging such as the snack bag (Fig. \ref{Fig_ComplexGrasp}(b)), the passive contact of the origami modules prevented crushing or surface deformation, while rigid objects like the yogurt cup or carton (Fig. \ref{Fig_ComplexGrasp}(a, c)) were firmly held under stable compression. The v-enveloping grasp naturally formed for irregular shapes such as the apple and egg (Fig. \ref{Fig_ComplexGrasp}(e, f)), providing robust and symmetric force closure with minimal slippage.
            

            Multi-object grasping reveals that the gripper is able to grasp a variety of objects simultaneously and with stability. In this situation, the grasp reconfigures such that the two origami module layers each hold one of the two objects, independently holding them stable. Depending on their position, spherical objects are either held in a firm parallel grasp as shown in Fig. \ref{Fig_ComplexGrasp}(i) or a half v-enveloping grasp which relies on gravity or the top of the gripper to maintain force closure, as seen in Fig. \ref{Fig_ComplexGrasp}(h, j). 

        \section{Discussion}
        The prevalent method of improving grasping performance has traditionally relied on increasing sensor density, improving perception, and implementing sophisticated control strategies. While effective, such methods often introduce significant system complexity and cost. 
        In contrast, this work presents a passive alternative based on deformable origami modules that embody mechanical intelligence within their structure and material. Despite being actuated by only a 1-DoF mechanism and manipulating without any sensors or feedback, the proposed gripper demonstrates the ability to stably grasp complex and varied geometries, maintain a predictable constant output, and even manipulate multiple objects simultaneously.        

        Experimental characterization validated that the origami modules generate constant output forces and torques within an effective deformation range for both materials tested. The TPU-based modules, offering higher bending torque and compression force, are more suitable for applications requiring firm yet compliant contact, such as handling irregular or dense objects. In contrast, the silicone-based modules exhibit superior linearity in both torque and compression output (Fig. \ref{Fig_CompressionTorque} and \ref{Fig_GraspPullOutTest}), making them ideal for grasping fragile or deformable items like snack bag. Moreover, the origami-based structure allows the mechanical characteristics to be reprogrammed through crease geometry and panel thickness, offering a simple and scalable solutions to existing soft gripper applications.
    
        Multi-object manipulation is a key feature of the proposed gripper, as Fig. \ref{Fig_MultiObjectGrasp} showcases proficiency in stably grasping a variety of object geometries, even without the use of v-enveloping grasping for the spherical objects. One current limitation is that the lower object must be smaller than the upper one to ensure selective placement, suggesting the need for future designs with variable module spacing or independently controlled fingers. Nonetheless, this functionality could provide substantial efficiency gains in industrial sorting, logistics, or packaging tasks, where rapid multi-object handling is beneficial. The simplicity of the mechanical actuation combined with intrinsic compliance provides an advantage over rigid or sensorized grippers.

        
        The gripper also performed robustly in handling daily-use objects of diverse shapes and materials (Fig. \ref{Fig_ComplexGrasp}).
        Spherical items such as the apple were effectively secured through an adaptive v-enveloping grasp, while cuboid and cylindrical items were accommodated via friction-based parallel grasps. A limitation of the four-finger grasp orientation is a narrow clearance width for objects extending beyond the grasp space, such as the banana (Fig. \ref{Fig_ComplexGrasp}(d)) or cuboid shapes (Fig. \ref{Fig_MultiObjectGrasp}). Future work will explore alternative finger orientations and modular configurations to expand grasp coverage and enhance adaptability across different object geometries.
        
		\section{Conclusion}
                In this paper, we presented a multi-finger hybrid gripper that integrates soft deformable origami modules to achieve passive, constant-force grasping of complex, fragile, and multiple objects simultaneously. The proposed gripper combines the simplicity of 1-DoF actuation with the mechanical intelligence of origami-inspired modules, enabling adaptive grasping without sensing. We detailed the mechanism and fabrication process, and analyzed the performance characteristics of its different materials in terms of constant force and torque behavior. Mechanical testing validated the gripper’s ability to maintain predictable grasping and pull-out forces, while experiments with various daily items demonstrated its versatility and robustness across diverse geometries. Furthermore, the gripper successfully performed multi-object grasping, achieving more efficient pick-and-place through adjustment of grasp size.

                Future work will focus on optimizing the durability deformation range of the origami modules, and exploring surface modifications such as sticky and textured interfaces to enhance frictional grip. Extending the multi-object manipulation capability to objects of identical size through layered offset configurations is also of interest. Finally, systematic studies of finger count, spacing, and orientation will be conducted to identify the optimal configurations for specific manipulation scenarios. Overall, this work demonstrates a promising direction for simple, sensor-free, and mechanically adaptive grippers, bridging the gap between soft robotics and practical multi-object manipulation.

		
		\bibliographystyle{IEEEtran}
		\bibliography{citationlist}

	\end{document}